\documentclass[10pt,twocolumn,letterpaper]{article}

\usepackage[pagenumbers]{cvpr} 

\usepackage{graphicx}
\usepackage{amsmath}
\usepackage{amssymb}
\usepackage{booktabs}
\usepackage{epigraph}
\usepackage{algorithm}
\usepackage{algpseudocode}
\usepackage{comment}
\usepackage{enumitem}
\setlist{nolistsep}
\usepackage{dblfloatfix}

\usepackage{hyperref}
\hypersetup{breaklinks,colorlinks}

\usepackage[capitalize]{cleveref}
\crefname{section}{Sec.}{Secs.}
\Crefname{section}{Section}{Sections}
\Crefname{table}{Table}{Tables}
\crefname{table}{Tab.}{Tabs.}

\newcommand{\RNum}[1]{\uppercase\expandafter{\romannumeral #1\relax}}

\def\cvprPaperID{10921} 
\def\confName{CVPR}
\def\confYear{2023}

\begin{document}

\title{FaceXHuBERT: Text-less Speech-driven E(X)pressive 3D Facial Animation Synthesis Using Self-Supervised Speech Representation Learning}

\author{Kazi Injamamul Haque\\
Utrecht University\\
{\tt\small k.i.haque@uu.nl}
\and
Zerrin Yumak\\
Utrecht University\\
{\tt\small z.yumak@uu.nl}
}
\maketitle

\begin{abstract}
This paper presents FaceXHuBERT, a text-less speech-driven 3D facial animation generation method that allows to capture personalized and subtle cues in speech (e.g. identity, emotion and hesitation). It is also very robust to background noise and can handle audio recorded in a variety of situations (e.g. multiple people speaking). Recent approaches employ end-to-end deep learning taking into account both audio and text as input to generate facial animation for the whole face. However, scarcity of publicly available expressive audio-3D facial animation datasets poses a major bottleneck. The resulting animations still have issues regarding accurate lip-synching, expressivity, person-specific information and generalizability. We effectively employ self-supervised pretrained HuBERT model in the training process that allows us to incorporate both lexical and non-lexical information in the audio without using a large lexicon. Additionally, guiding the training with a binary emotion condition and speaker identity distinguishes the tiniest subtle facial motion. We carried out extensive objective and subjective evaluation in comparison to ground-truth and state-of-the-art work. A perceptual user study demonstrates that our approach produces superior results with respect to the realism of the animation 78\% of the time in comparison to the state-of-the-art. In addition, our method is 4 times faster eliminating the use of complex sequential models such as transformers. We strongly recommend watching the supplementary video before reading the paper. We also provide the implementation and evaluation codes with a GitHub repository link.
\noindent\footnotesize\url{https://github.com/galib360/FaceXHuBERT}
\end{abstract}

\section{Introduction}
\label{sec:intro}
\begin{figure}[t]
  \centering
   \includegraphics[width=0.95\linewidth]{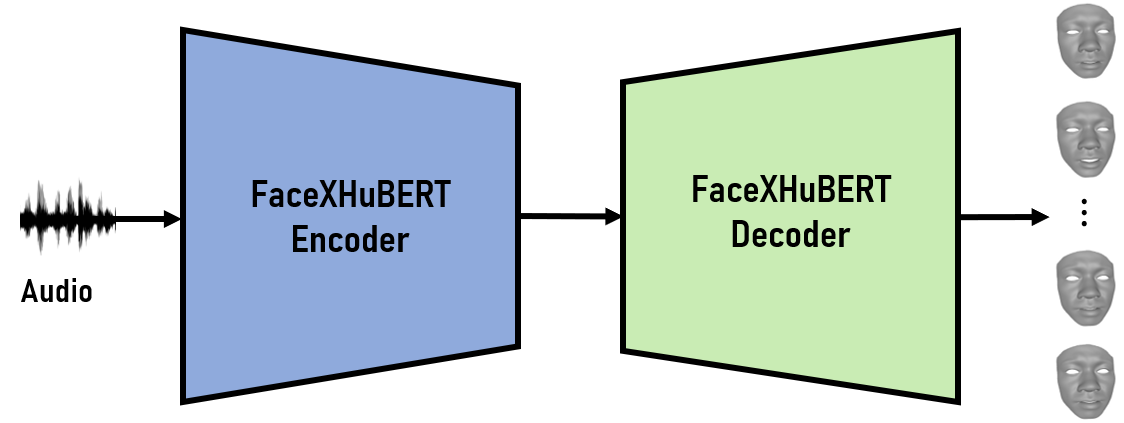}
   \caption{\textbf{FaceXHuBERT:} An end-to-end encoder-decoder architecture that encodes audios using self-supervised pretrained speech model HuBERT and decodes to vertex displacements using GRU followed by a fully connected linear layer that produces 3D facial animation as 3D mesh sequences.}
   \label{fig:model_simple}
   \vspace{-5mm}
\end{figure}

Speech-driven 3D facial animation is a growing yet challenging research area with applications to games, VR/AR and film production. Conversational virtual humans with social and emotional interaction capabilities are used in a range of applications such as chatbots for customer service and marketing, simulations for education and healthcare and remote communication. Facial expressions are the first point of attention in conversational communication and humans are very receptive to subtle nuances in facial animation which is explained by the uncanny valley theory \cite{Mori2012}. 
 
Typically, facial animation workflows rely on professional technical artists using blendshape facial animation \cite{Blendshape} or performance capture aiming to mitigate most of the labor intensive work \cite{Dynamixyz, faceware, di4d}. However, as these characters take place in more interactive applications, the demand to automatically generate their behavior on-the-fly increases. Research on facial animation focuses on 2D talking faces \cite{talkingFace4, Lu2021, talkingFace8, Perov2020}, 3D facial animation constructed from 2D images and videos \cite{DECA:Siggraph2021, Ma2021, EMOCA:CVPR:2021, Zhang2022} and 3D speech-driven facial animation \cite{Taylor2017, TeroKarras, MeshTalk, visemenet2018, VOCA2019, faceformer2022, Voice2FaceEA}. In this paper, we propose a novel approach for 3D speech-driven facial animation. 

3D speech-driven facial animation is either based on phoneme-based approaches using procedural algorithms \cite{jali, Charalambous2019} or data-driven approaches using machine learning \cite{Taylor2012}, motion graphs \cite{Cao2005} and deep learning \cite{Taylor2017, visemenet2018}. The former requires explicit definition of co-articulation rules and requires manual work. While the latter aims to eliminate that by learning speech-animation parameters mapping from data, it still relies on intermediary representations of speech units. Recent approaches on 3D speech animation synthesis effectively employ end-to-end deep learning models \cite{TeroKarras, MeshTalk, visemenet2018, VOCA2019, faceformer2022} eliminating the need for intermediary representations. Large speech and language models \cite{HuBERT, GPT3} open the way towards more realistic speech-driven facial animation. However, the lack of 3D facial animation data matching audio and text poses a major bottleneck and current models cannot generalize to arbitrary speech input. Most recently, Fan et al. \cite{faceformer2022} suggested a self-supervised speech representation learning method using transformers to mitigate these issues. However, the method cannot handle expressive animations, the results are still far from capturing personalized and subtle cues in speech and the method is computationally very expensive to train. 

Our work improves on this work and proposes FaceXHuBERT, a text-less speech-driven expressive 3D facial animation generation method using self-supervised speech representation learning. In our proposed encoder-decoder network (see \cref{fig:model_simple}), we effectively employ self-supervised pretrained HuBERT model to incorporate and encode both lexical and non-lexical information without using a large lexicon and speech-3D data pairs allowing it to generalize to any speech input.  Our method is simple and efficient eliminating the use of complex sequential models such as transformers and instead uses a decoder based on Gated Recurrent Units. 
The main contributions of our work are: 

\begin{itemize}
    \item \textbf{An efficient text-less speech-driven expressive 3D facial animation method using self-supervised speech representation learning}. FaceXHuBERT produces expressive and realistic animations in an efficient way using a HuBERT-based encoder and GRU-based decoder without the use of a large lexicon and using only audio input. Additionally, guiding the training with an emotion condition and speaker identity distinguishes the tiniest subtle facial motions. The results show that our method produces more realistic results in a more efficient manner (i.e. 4 times faster in comparison to a vanilla transformer and almost 3 times faster in comparison to state-of-the-art \cite{faceformer2022}).
    \item \textbf{Proof of self-supervised pretrained speech model HuBERT \cite{HuBERT} for the downstream task of expressive 3D facial animation synthesis}. Our method produces accurate lip-sync as well as allows to capture personalized and subtle cues in speech (e.g. identity, emotion and hesitation). It is also very robust to background noise and can handle audio recorded in a variety of situations (e.g. multiple people speaking, background noise, laughter, lip-smacking).
    \item \textbf{Extensive objective and subjective analyses}. We compared our method to state-of-the-art and ground-truth as well as made comparisons between various seq-to-seq neural network architectures using 3D vertex error as an objective metric. Subjective analysis includes qualitative generalizability analysis in terms of different languages, text-to-speech, noise, low-quality audio input and single subject training. We also conducted several perceptual user studies. Our results demonstrate that our approach produces superior results with respect to the realism of the animation 78\% of the time in comparison to the state-of-the-art \cite{faceformer2022}.
\end{itemize}

\section{Related Work}

Extensive research has been conducted in the domain of automatic facial expression analysis and synthesis in the 2D pixel domain for the purpose of detecting expressions \cite{AU1, AU2, AU3}, for generating audio-driven talking faces \cite{talkingFace2, talkingFace3, talkingFace4, Lu2021, talkingFace8} or for video-based facial re-enactment/face swapping \cite{Suwajanakorn2017, Nirkin2019, Perov2020}. 

The approaches for 3D facial animation synthesis can be classified into video-driven and audio-driven facial animation. While the former focuses on transferring facial animation from 2D videos to 3D faces, the latter maps speech (audio and text) to 3D facial animation parameters. Earlier works on video-driven facial animation focused on optimization-based 3D facial performance capture \cite{Bouaziz2013, Cao2014, Ichim2015}, while recent works use deep learning \cite{RingNet:CVPR:2019, DECA:Siggraph2021, Ma2021, EMOCA:CVPR:2021, moser2021}. For an extensive survey on 3D face reconstruction, tracking and morphable models, we refer to \cite{MORALES2021100400, Zollhofer2018, Egger2020}. Some methods use retargeting algorithms to convert facial expressions from one 3D mesh to the other \cite{Ribera:2017, Chandran2022} or from 2D images to 3D faces \cite{Yang2020, Zhang2022}. Finally, there is a group of research focusing on physics-based animation of faces \cite{Ichim2017, Barrielle2018}. In our work, we focus on 3D speech-driven facial animation using deep learning.
\vspace{-1em}

\paragraph{3D Speech-Driven Facial Animation}

3D speech-driven facial animation typically uses phoneme-based procedural approaches \cite{jali, Charalambous2019}. Although these methods come with the advantage of animation control and easy integration to artist-friendly pipelines, they are not fully automatic and require defining explicit rules for co-articulation. Another line of research uses machine learning \cite{Taylor2012} or graph-based approaches \cite{Cao2005} to learn speech-animation mappings from data. These methods rely on blending between speech units and cannot capture the complexity of the dynamics of visual speech \cite{Taylor2017}. They rather focus on the lower face and are not robust to emotion and style variations. Recent approaches on 3D speech animation synthesis effectively employ deep learning models \cite{Taylor2017, TeroKarras, MeshTalk, visemenet2018, VOCA2019, faceformer2022, Voice2FaceEA}. Taylor \etal\cite{Taylor2017} proposes a sliding window approach instead of an RNN focusing on capturing neighborhoods of context and coarticulation effects. VisemeNet \cite{visemenet2018} builds upon the viseme-based JALI\cite{jali} model and combines this with an LSTM-based neural network. However, these two methods \cite{Taylor2017, visemenet2018} still rely on intermediary representations of phonemes and they focus on the mouth movement. Most previous works do not include automatic tongue animation except \cite{Voice2FaceEA}. Collecting large-scale datasets using professional performance capture workflows is expensive and time consuming but the resulting faces are highly realistic. To elevate this disadvantage, some methods use 3D automatic face reconstruction methods from in-the-wild videos, which are especially useful in situations where professional performance capture systems are not available, e.g. dyadic speech-driven facial animation \cite{jonell2020letsfaceit, ng2022learning2listen}. However, these methods are prone to 3D reconstruction errors and cannot generate results that are as realistic as the former.

Closest to our work are Karras \etal\cite{TeroKarras}, Cudeiro \etal \cite{VOCA2019}, Richard \etal\cite{MeshTalk}, Fan \etal\cite{Fan21JA, faceformer2022}. Karras \etal\cite{TeroKarras} proposes an end-to-end convolutional neural network that learns a mapping from input waveforms to the 3D vertex coordinates of a face model. They aim to resolve the ambiguity in mapping between audio and face by introducing an additional emotion component to the network, which is learned from data. However, the method is not trained on multiple speakers and cannot handle identity variations and requires a longer-term audio context to infer the emotional state. Instead, Cudeiro \etal \cite{VOCA2019} presents the audio-driven facial animation method VOCA that generalizes to new speakers using a training dataset with 12 subjects eliminating the need for retargeting. However, VOCA fails to realistically synthesize upper face motion and does not include emotional variations. Similar to VOCA, Richard \etal\cite{MeshTalk} aims for audio-driven animation that can capture variations in multiple speakers including a much larger dataset of 250 subjects. They address the problem of lack of upper face motions using a categorical latent space that disentangles audio-correlated and audio-uncorrelated information based on a cross-modality loss. Fan \etal\cite{Fan21JA} proposes an audio and text-driven facial animation method that incorporates the large language model GPT-2\cite{gpt2} to encode the textual information. The authors found that combined audio and text input yielded better results than audio-only or text-only model. However, the results still have problems regarding accurate lip-sync. Most closely related to our work is FaceFormer\cite{faceformer2022} which uses a self-supervised pretrained speech model that addresses the scarcity of available data in existing audio-visual datasets. The model produces superior results in comparison to Cudeiro \etal \cite{VOCA2019} and Richard \etal\cite{MeshTalk} using a modified version of transformers to handle longer sequences of data. However, none of these methods can handle arbitrary variations in speech input while producing accurate lower and upper facial animation for multiple identities and emotions.

\section{Problem Formulation}
We formalize the task of audio-driven 3D facial animation as a generic sequence modeling (seq2seq) problem in which the input sequence is a raw audio waveform whereas the output sequence is a 3D face mesh sequence (i.e. 4D scan). Hence, the problem can be formalized as follows:

Given audio $A$ and ground-truth 3D mesh sequence Y$ = (y_1, y_2, y_3,..., y_{T_Y})$, $T_Y$ is the total number of available visual frames or 3D scanned frames in the sequence. Therefore, one sequence of Y is a $(T_Y, V)$ dimensional matrix where $V$ denotes the number of 3D vertices in the mesh topology. On the input side, since audio stream $A$ is a continuous data stream, with the help of an encoder, we encode the continuous audio into a discrete representation $X = (x_1, x_2, x_3,..., x_{T_X})$ where $X$ is a $(T_X, B)$ dimensional matrix and $T_X$ and $B$ are the discreet time-steps and the encoded representation respectively. 

The goal is to train an end-to-end architecture to learn the mapping between $A$ (together with additional conditions) and Y to generate $\hat{Y} = (\hat{y}_1, \hat{y}_2,\hat{y}_3,..., \hat{y}_{T_y})$ so that the $\hat{Y}$ best approximates Y. 

\section{Proposed Approach}
\label{sec:proposedapproach}
\begin{figure*}
  \centering
  \includegraphics[width=0.98\linewidth]{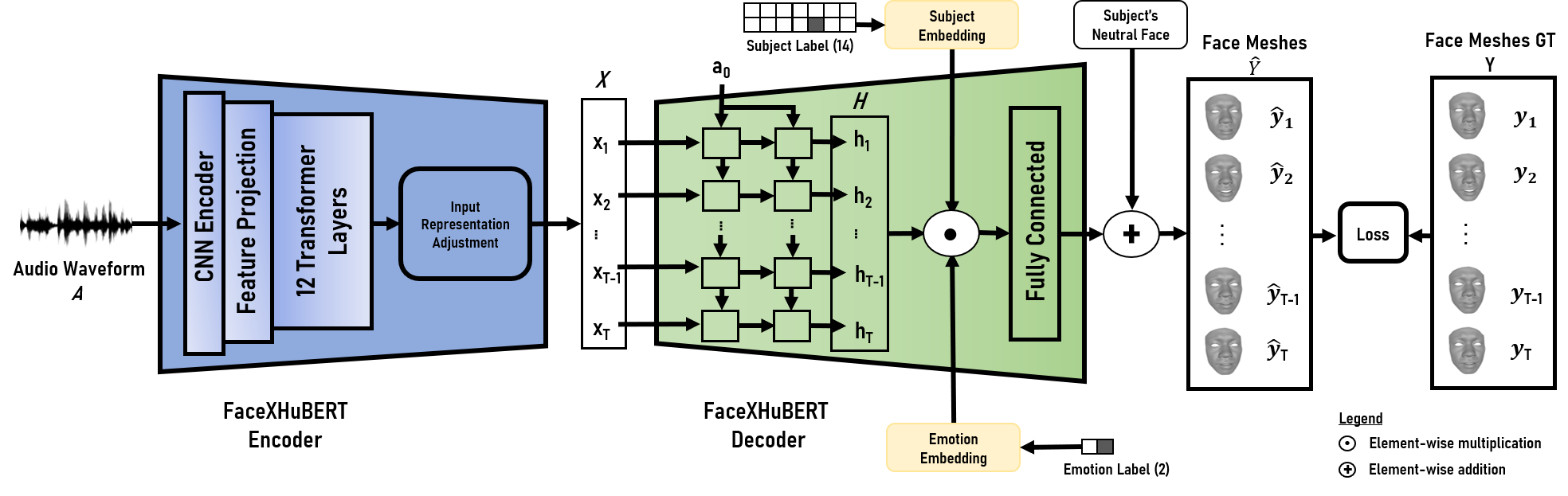}
  \caption{\textbf{FaceXHuBERT:} The encoder encodes the audio waveform $A$ and produces discreet frame level embedding. The Input Representation Adjustment module in the encoder adjusts the encoded information with the output 4D scan data and produces $X$ such that $T_X = T_Y = T$. The Decoder takes in $X$ and with the help of the 2-layered 256 hidden sized GRU, produces the hidden representation $H$. Additional conditions such as the speaker identity and emotion label are embedded and multiplied with $H$ before the hidden representation is decoded into vertex displacement values and added to the corresponding subject's neutral face to produce the animation output $\hat{Y}$. The loss function is computed based on $\hat{Y}$ and ground-truth (GT), Y.}
  \label{fig:ProposedApproach}
\end{figure*}

We present FaceXHuBERT, an end-to-end encoder-decoder neural network architecture. Our model uses the pretrained HuBERT speech model as the audio encoder while for the decoder, we use Gated Recurrent Unit (GRU) \cite{GRU}. \cref{fig:ProposedApproach} shows the overall architecture of our proposed approach. The encoder encodes the continuous audio information into discreet time-step representations and adjusts the representations so that the time-steps match with that of the face scan data. The decoder incorporates the emotion and subject identity information, consists of a 2-layered GRU with hidden size 256 followed by a fully connected linear layer. The decoder regresses vertex displacements and adds to the subject's template mesh to generate the predicted mesh sequence.  \cref{alg:training} in the supplementary material depicts the overall steps of the proposed network. In the next two subsections, we describe the details of the FaceXHuBERT Encoder and Decoder.

\subsection{FaceXHuBERT Encoder}
\label{sec:enc}
Our proposed method effectively adopts the state-of-the-art self-supervised pretrained speech model HuBERT in the encoder for the downstream task of 3D facial animation generation. Since it is able to learn and produce high quality discreet hidden representations of continuous audio streams combining both acoustic and language information, the authors of HuBERT recommend to consider using HuBERT pretrained representations for a variety of downstream tasks \cite{HuBERT}. HuBERT architecture introduces a BERT-like \cite{BERT} masked language modelling encoder for the transformer layers. It introduces a simple cross-entropy loss for predicting masked units in contrast to its predecessor Wave2Vec 2.0's \cite{Wav2Vec2} complex contrastive loss. In addition, unlike Wave2Vec 2.0, HuBERT is trained with multiple iterations. During the first iteration, HuBERT uses unsupervised simple k-means clustering for acoustic unit discovery to facilitate the self-supervised masked language modeling learning that takes place in the second iteration. In the second iteration, the training is done on the discovered discrete hidden units with a predictive loss on the masked regions only, forcing the model to learn a combined acoustic and language model using a BERT-like encoder, hence the name H(idden)-u(nit)-BERT. The model is trained on 960 hours of unlabeled speech data \cite{librispeech}  which contains English recordings of copyright-free audiobooks by volunteers from the internet. The authors claim that this approach is the first big step towards text-less Natural Language Processing (NLP). For detailed explanation of how different HuBERT models were trained and for comparison to previous work, we refer to the original paper \cite{HuBERT}. 

The FaceXHuBERT Encoder is composed of a CNN encoder that discretizes the continuous audio data into 512 dimensional representations. Feature Projection layer projects the 512 dimensional representation into 768 dimensional representation, a positional convolution embedding layer and 12 transformer layers to capture the contextual information in the sequence. In our approach, we adopt the ``base" HuBERT model with 95M parameters which produces 768 dimensional embedding at the last hidden state. We initialize the pretrained weights and freeze the model parameters including the CNN feature encoder layer, feature projection layer and the first two transformer layers. The last ten transformer layers are kept unfrozen and remain trainable. HuBERT generates a feature sequence in 20ms windows (i.e. 50 fps). Therefore, in our architecture, we encode a one second of audio into 50 frames with 768 dimensional embeddings. For example, for a training data with 4 seconds of audio stream, the output from the encoder will be $(4\times50, 768) = (200, 768)$ dimensional matrix. 
\vspace{-1em}
\paragraph{Input Representation Adjustment}
This module adjusts the input representation and output representations and do not contain any trainable parameters. This is devised to ensure the one-to-one frame level relationship between decoder input $X$ and output $Y$ such that $T_X = T_Y = T$. This function is generically devised in such a way that it can handle any input-output frequency pair. More details on this can be found in the supplementary material.

\subsection{FaceXHuBERT Decoder}
For the FaceXHuBERT Decoder, we use a Gated Recurrent Unit (GRU) instead of a complex transformer model. Our extensive analysis shows that, combined with the HuBERT encoder, GRU-based decoder produces realistic results in a more efficient manner. Our GRU-based decoder consists of 2 layers with hidden unit size of 256, followed by one fully connected linear layer that maps the last hidden state to vertex displacement values of the 3D vertices of the face. It represents the faces in terms of their displacement values with respect to the neutral template vertices of a given subject. Between the GRU and the fully connected layer, we add the additional conditions and fuse them with the hidden state representation with element-wise multiplication. The additional conditions are (i) subject identity and (ii) emotion (neutral or expressive in our experiments). We defined the training subjects and the emotion label as one-hot vectors and linearly embed them with two separate 256 dimensional vectors to facilitate the element-wise multiplications. 

\cref{eq:r-gate}, \cref{eq:new_candidate}, \cref{eq:u-gate} and \cref{eq:new_a} show the core computations of our decoder's forward propagation.

\begin{equation}
  \Gamma_r = \sigma(W_r^{l} [a_{t-1}^{l}, a_t^{l}] + b_r^{l})
  \label{eq:r-gate}
\end{equation}
\vspace{-1.5em}
\begin{equation}
  \Tilde{a}_t^{l} = tanh (W_a^{l} [\Gamma_r \odot a_{t-1}^{l}, a_t^{l}] + b_a^{l})
  \label{eq:new_candidate}
\end{equation}
\vspace{-1.5em}
\begin{equation}
  \Gamma_u = \sigma(W_u^{l} [a_{t-1}^{l}, a_t^{l}] + b_u^{l})
  \label{eq:u-gate}
\end{equation}
\vspace{-1.5em}
\begin{equation}
  a_t^{l} = \Gamma_u \odot \Tilde{a}_t^{l} + (1-\Gamma_u) \odot a_{t-1}^{l}
  \label{eq:new_a}
\end{equation}

The subscript $t$ denotes the frame number or time-step in the sequence whereas the superscript $l$ denotes the hidden layer where $l = [1,2]$ (i.e. two hidden layers). When $l=1$, the activation values $a_t^{l}$ (not $a_{t-1}^{l}$) in \cref{eq:r-gate}, \cref{eq:new_candidate} and \cref{eq:u-gate} take the input value $x_t$, hence, $a_t^{1}=x_t$. When $l=2$, the activation values $a_t^{2} = h_t$, each being 256 dimensional hidden unit. The second GRU layer produces the hidden units $H = [h_1, h_2, h_3, ..., h_T]$ for a sequence with $T$ frames. Furthermore, we initialize $a_0 = \Vec{0}$ to start the training.  

\cref{eq:S}, \cref{eq:E}, \cref{eq:h_tilde} and \cref{eq:Y_hat} show how the subject identity and emotion conditions are incorporated into the network.

\begin{equation}
  S = W_S \cdot [SubjectOneHot] + b_S
  \label{eq:S}
\end{equation}
\vspace{-1.5em}
\begin{equation}
  E = W_E \cdot [EmotionOneHot] + b_E
  \label{eq:E}
\end{equation}
\vspace{-1.5em}
\begin{equation}
  \Tilde{H} = H \odot S \odot E
  \label{eq:h_tilde}
\end{equation}
\vspace{-1.5em}
\begin{equation}
  \hat{Y} = (W_{\hat{Y}} \cdot \Tilde{H} + b_{\hat{Y}}) \oplus [NeutralFace] 
  \label{eq:Y_hat}
\end{equation}

The subject label and the emotion label are represented as one-hot vectors that we linearly embed to 256 dimensional $S$ and $E$ vectors using \cref{eq:S} and \cref{eq:E} respectively. In \cref{eq:h_tilde}, the hidden representation $H$ with dimensions $(T,256)$ is multiplied in an element-wise manner with both the embedding vectors of the given subject, $S$ (i.e. training subjects) and the style of the given sequence, $E$ (i.e. neutral or expressive). Finally, in \cref{eq:Y_hat}, the fully connected linear layer decodes the hidden representation into $(T,V)$ dimensional vertex displacement values which is then added to the respective subject's neutral face vertex values, where $[NeutralFace]$ is a $(1,V)$ dimensional data.

\begin{figure*}
  \centering
\begin{subfigure}{0.15\linewidth}
    \includegraphics[width=\linewidth,trim={3.3cm 3.3cm 3.3cm 3.3cm},clip]{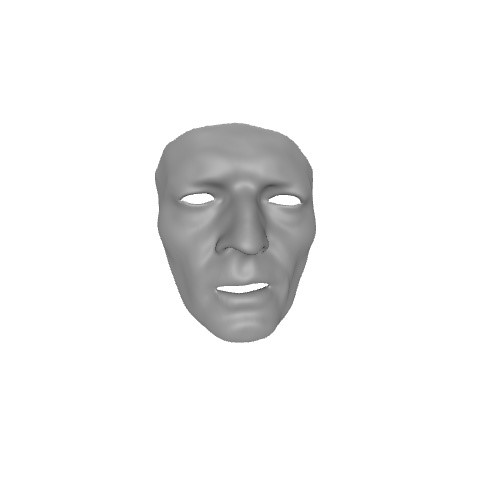}
    \caption{Neutrally generated animation}
    \label{fig:emoneutral-a}
  \end{subfigure}
  \hfill
  \begin{subfigure}{0.15\linewidth}
    \includegraphics[width=\linewidth,trim={3.3cm 3.3cm 3.3cm 3.3cm},clip]{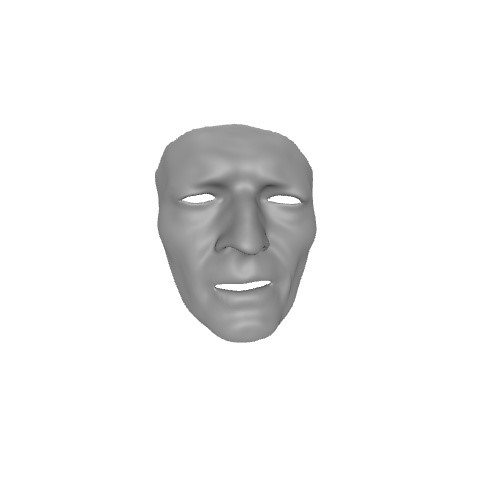}
    \caption{Expressively generated animation}
    \label{fig:emoneutral-b}
  \end{subfigure}
  \hfill
  \begin{subfigure}{0.15\linewidth}
    \includegraphics[width=\linewidth,trim={3.3cm 3.3cm 3.3cm 3.3cm},clip]{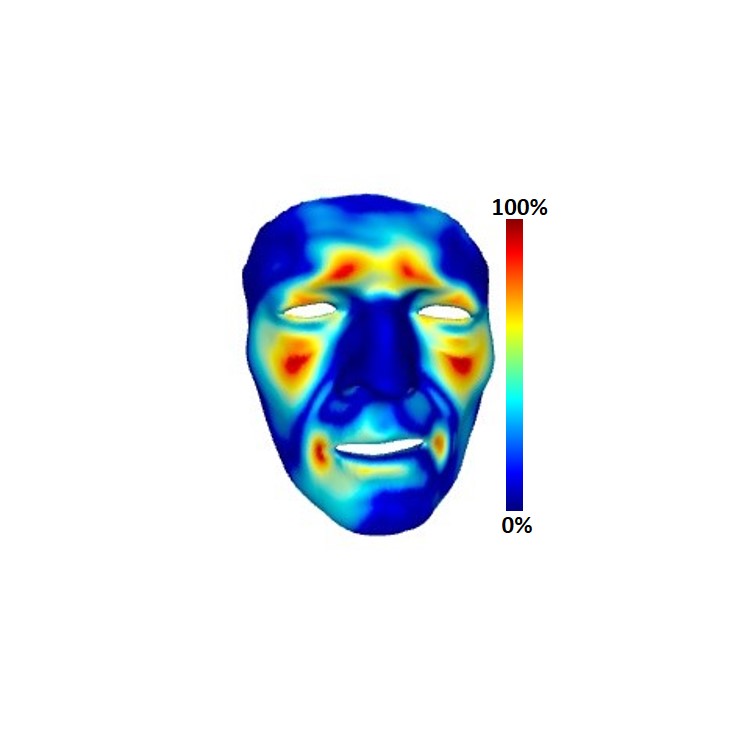}
    \caption{Difference between (a) \& (b)}
    \label{fig:emoneutral-c}
  \end{subfigure}
  \hfill\begin{subfigure}{0.15\linewidth}
    \includegraphics[width=\linewidth,trim={3.2cm 3.2cm 3.2cm 3.2cm},clip]{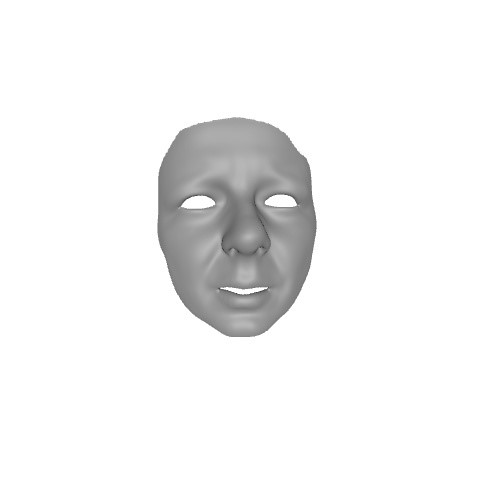}
    \caption{Neutrally generated animation}
    \label{fig:emoneutral-d}
  \end{subfigure}
  \hfill\begin{subfigure}{0.15\linewidth}
    \includegraphics[width=\linewidth,trim={3.2cm 3.2cm 3.2cm 3.2cm},clip]{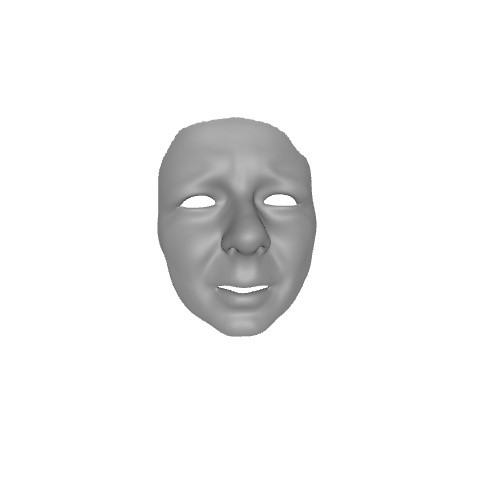}
    \caption{Expressively generated animation}
    \label{fig:emoneutral-e}
  \end{subfigure}
  \hfill\begin{subfigure}{0.15\linewidth}
    \includegraphics[width=\linewidth,trim={3.2cm 3.2cm 3.2cm 3.2cm},clip]{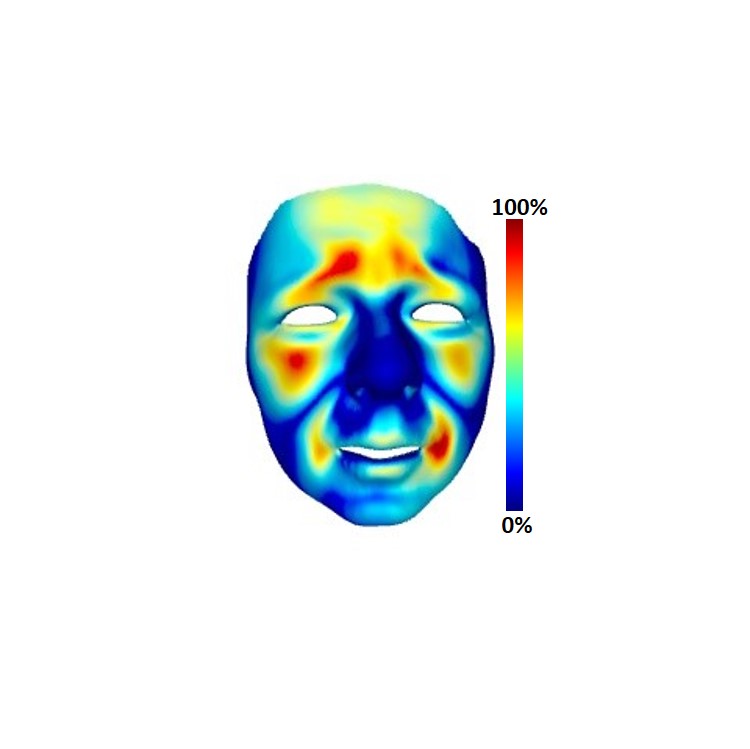}
    \caption{Difference between (d) \& (e)}
    \label{fig:emoneutral-f}
  \end{subfigure}
  \caption{\textbf{Effect of the emotion label during inference:} our approach can generate facial animations that are style controllable by a binary emotion label. Given two in-the-wild audio signal examples, \cref{fig:emoneutral-a,fig:emoneutral-b,fig:emoneutral-c} correspond to the male example whereas \cref{fig:emoneutral-f,fig:emoneutral-e,fig:emoneutral-f} correspond to the female example. \cref{fig:emoneutral-a,fig:emoneutral-d} are generated neutrally whereas \cref{fig:emoneutral-b,fig:emoneutral-e} are generated expressively. \cref{fig:emoneutral-c,fig:emoneutral-f} show the colorized differences based on per-vertex distances between neutrally and expressively generated face meshes where extreme red depicts 100\% of the computed distance and extreme blue depicts 0\% of the computed distance. It is evident that the emotion signal effects the facial regions that are uncorrelated with speech.}
  \label{fig:emoeffect}
\end{figure*}

\section{Experimental Setup}

\subsection{Dataset and Pre-processing}
For our experiments, we used the BIWI\cite{eth_biwi_00760} dataset which is the only dataset that is available publicly upon request with both emotion and identity labels. VOCASET\cite{VOCASET} and Multiface\cite{multiface} datasets allow incorporating identities but not emotions. BIWI contains synchronized audio-4D scan pairs of 14 human subjects uttering 40 phonetically balanced English sentences twice: first neutrally, second with emotional expressions. Therefore, the dataset contains $(14\times40\times2)=1120$ sequences of audio-4D scan pairs. In reality, due to some missing sequences from 4D scans, there are in total 1088 audio-4D pairs. The sequences in the dataset are 4.39 seconds in duration on average including both neutral and emotional sequences. Furthermore, on average, neutral sentences are 4.28 seconds long whereas the emotional sentences are 8.90 seconds long. To ensure efficient training, we pre-process the data by scaling the 3D vertices to have a uniform range of values for all three coordinates across the dataset (e.g. [-0.5,0.5]). The data is captured at 25fps with 23370 3D vertices in the mesh topology. The pre-processed dataset is then split in 90\% train, 5\% validation and 5\% test sets. More details on dataset pre-processing can be found in the supplementary material.

\subsection{Model Experiments}
\label{sec:experiments}
We experimented with different seq-to-seq neural network architectures including a vanilla RNN, LSTM, GRU and Transformer as FaceXHuBERT Decoder. In this section, we will briefly discuss about the model experiments and their results.

\noindent\textbf{HuBERT-RNN} model is devised by using RNN with the 2 hidden layers and with hidden size 256. The resulting 3D facial animations are acceptable in terms of visual quality and lip-sync. However, it does not generalize well for longer in-the-wild audios. The resultant animations are not smooth, noisy and show visual artefacts.

\noindent\textbf{HuBERT-LSTM} model uses LSTM with 2 hidden layers and with hidden size 256. The resulting animations start to move in the very beginning but immediately becomes static and maintains the stagnant pose throughout the remaining frames of the sequence. As mentioned in \cite{fragkiadaki}, LSTM dominates for short term motion generation yet soon converges to a mean pose.

\noindent\textbf{HuBERT-Transformer} uses a vanilla transformer as decoder. We trained using both teacher forcing and autoregressive approaches. The former yields static facial pose throughout the animation sequence while the latter produces meaningful animations. However, due to the network complexity, the training time is 4 times slower when compared to our proposed approach. Additionally, the transformer based approach is constrained by a pre-defined maximum length of input sequence.

\noindent\textbf{HuBERT-GRU} network architecture of this model is our proposed model described in \cref{sec:proposedapproach}. Using GRU in the vertex decoder yields audio coherent animations for both test-set sequences as well as arbitrarily long in-the-wild audio sequences recorded in variety of situations. This model produces the most realistic and expressive results. Extensive evaluations can be found in the following section.

\noindent\textbf{Training Details and Tools} All the models in our experiments were trained for 100 epochs on an HP ZBook Fury G7 laptop with an Intel Core i7-10850H 2.7 GHz (12 cores) CPU, 32GB RAM and Nvidia Quadro RTX 3000 6GB VRAM. We optimized on Huber Loss function \cite{huberloss} and used Adam optimizer \cite{adam} during the training. The dropout value of the recurrent units was set to $0.3$. We used PyTorch \cite{PyTorch} for implementing the models. Meshlab \cite{MeshLab} and PyMeshLab \cite{pymeshlab} were used to compute mean face vertex error for quantitative evaluation. Trimesh \cite{TrimeshLib} together with OpenCV \cite{opencv_library} and ffmpeg \cite{ffmpeg} were used for rendering and visualizing. The entire codebase for training, evaluating and visualizing can be found in the supplementary material.

\section{Evaluation}
Evaluation of the 3D facial animation generation is challenging and there is no unanimously accepted single objective metric in the literature. Therefore, in addition to extensive quantitative evaluations using mean face vertex error as an objective metric, we also conducted several qualitative evaluations in terms of visual assessments. Furthermore, since facial animation is perceptual, we conducted user studies to compare our approach with ground-truth as well as the state-of-the-art. Because FaceFormer proved to be superior than the other state-of-the-art works, we compared our approach only with FaceFormer\cite{faceformer2022}. 
\begingroup
\setlength{\tabcolsep}{4pt} 
\renewcommand{\arraystretch}{0.6} 
\begin{table}[t]
  \centering
  \begin{tabular}{c c c c}
    \toprule
    Model & Mean Face & Training\\
    Type & Vertex Error\textsuperscript{\textcolor{red}{5}} & Time (h)\\
    \midrule

    HuBERT-RNN & $5.06$ & $\approx5.33$\\
    HuBERT-LSTM &  $5.98$ & $\approx5.55$\\
    HuBERT-Transformer\textsuperscript{\textcolor{red}{1}} & $8.74$ & $\approx5.27$\\
    HuBERT-Transformer\textsuperscript{\textcolor{red}{2}} & $5.13$ & $\approx25.00$\\
    FaceFormer\textsuperscript{\textcolor{red}{3}} & $5.95$ & $\approx6.38$\\ 
    \textbf{FaceXHuBERT}\textsuperscript{\textcolor{red}{3}} & $\mathbf{5.45}$ & $\mathbf{\approx1.52}$\\
    FaceFormer\textsuperscript{\textcolor{red}{4}} & $6.36$ & $\approx16.11$\\
    \textbf{FaceXHuBERT}\textsuperscript{\textcolor{red}{4}} & $\mathbf{4.80}$ & $\mathbf{\approx5.10}$\\
    FaceXHuBERT- w/o emo & $4.93$ & $\approx5.69$\\
    HuBERT-FaceFormer & $4.96$ & $\approx19.58$\\
    \bottomrule
    \multicolumn{3}{l}{\footnotesize \textsuperscript{\textcolor{red}{1}} Teacher-forcing scheme. Produces static animations.}\\
    \multicolumn{3}{l}{\footnotesize \textsuperscript{\textcolor{red}{2}} Autoregressive scheme.}\\
    \multicolumn{3}{l}{\footnotesize \textsuperscript{\textcolor{red}{3}} Model trained on only 6 subjects for fair comparison.}\\
    \multicolumn{3}{l}{\footnotesize \textsuperscript{\textcolor{red}{4}} Model trained on the whole dataset (i.e. all 14 subjects). FaceFormer }\\
    \multicolumn{3}{l}{\footnotesize does not produce good results when trained on the whole dataset.}\\
    \multicolumn{3}{l}{\footnotesize \textsuperscript{\textcolor{red}{5}} In millimeters (mm) when the face resides in a 1$m^{3}$ bounding box.}
  \end{tabular}
  \caption{Objective evaluation results of the experiments and trained models. Our approach not only produces the minimum Mean Face Vertex Error on test-set sequences but also reduces the training time significantly. FaceXHuBERT is more than 4 times faster than transformer based architecture and almost 3 times faster than the state-of-the-art.}
  \label{tab:Obj_evaluation}
  \vspace{-5mm}
\end{table}
\endgroup
\subsection{Quantitative and Qualitative Evaluation}
\label{sec:eval}
We measure and compare our proposed methodology quantitatively based on mean face vertex error. For each test-set sequence we take the vertex distance values of the predicted data with respect to the ground-truth reference for all the frames. We take the arithmetic mean of the calculated differences across the frames for all the test sequences and compute the average value to get the ``Mean Face Vertex Error". \cref{tab:Obj_evaluation} reports this error value together with corresponding training time for each of the models we presented in \cref{sec:experiments}. In addition, we also compared our approach with the state-of-the-art FaceFormer model. It is to be noted that FaceFormer fails to generate meaningful animations when trained on the whole BIWI dataset (i.e. all 14 subjects with all available sequences). Therefore, for a fair comparison, in addition to training on the whole dataset, we trained FaceFormer and FaceXHuBERT with the FaceFormer's suggested training including only 6 training subjects. The proposed FaceXHuBERT yields the minimum mean face vertex error while reducing the network complexity and training time significantly. Additionally, FaceFormer trains well on the whole dataset only when we replace their Wav2Vec2.0-based audio encoder with FaceXHuBERT encoder (HuBERT-FaceFormer in \cref{tab:Obj_evaluation}). However, the mean face vertex error is higher and the training is almost 4 times slower in comparison to ours. Besides, we trained our approach without incorporating the emotion label (FaceXHuBERT-w/o emo in \cref{tab:Obj_evaluation}) similar to how FaceFormer is trained.

\begin{figure}[t]
  \centering
\begin{subfigure}{0.24\linewidth}
    \includegraphics[width=\linewidth,trim={7.0cm 3.5cm 7.0cm 3.0cm},clip]{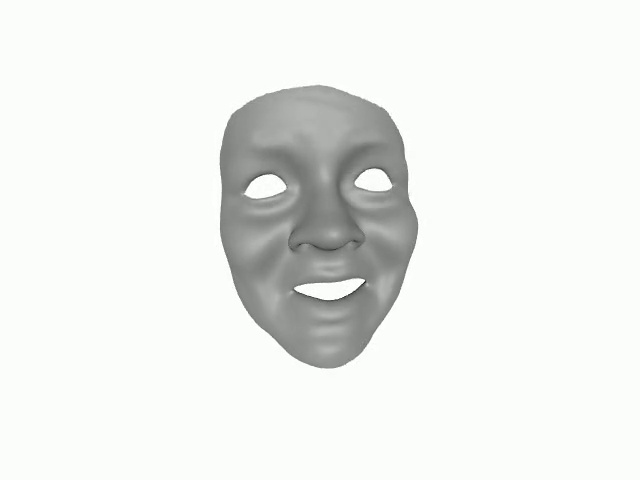}
    \caption{GT}
    \label{fig:ff_ours_gt}
  \end{subfigure}
  \begin{subfigure}{0.24\linewidth}
    \includegraphics[width=\linewidth,trim={7.0cm 3.5cm 7.0cm 3.0cm},clip]{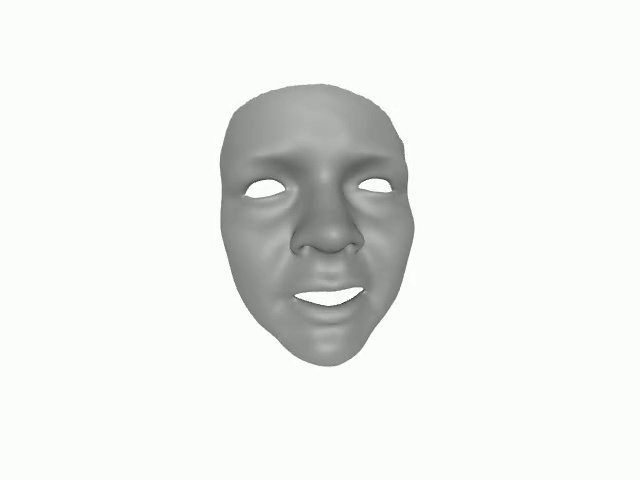}
    \caption{FaceFormer}
    \label{fig:ff_ours_ff}
  \end{subfigure}
  \begin{subfigure}{0.24\linewidth}
    \includegraphics[width=\linewidth,trim={7.0cm 3.5cm 7.0cm 3.0cm},clip]{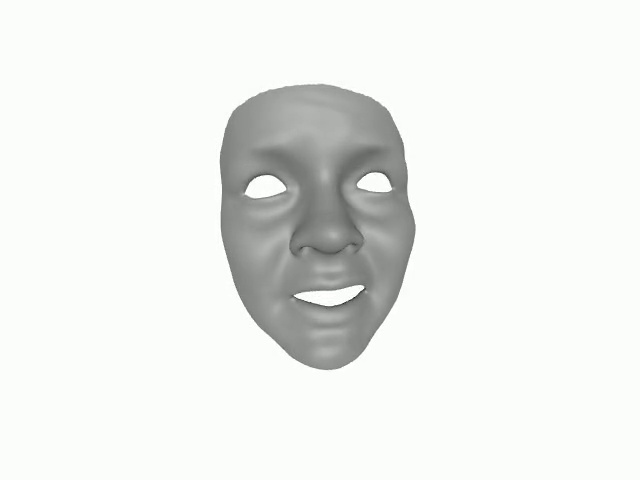}
    \caption{Ours}
    \label{fig:ff_ours_ours}
  \end{subfigure}
    \begin{subfigure}{0.24\linewidth}
    \includegraphics[width=\linewidth,trim={7.0cm 3.5cm 7.0cm 3.0cm},clip]{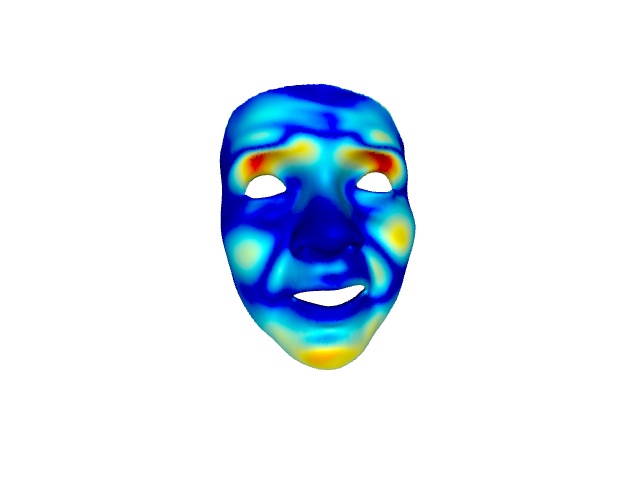}
    \caption{Difference}
    \label{fig:ff_ours_diff}
  \end{subfigure}
  \caption{Qualitative comparison of expressiveness between FaceFormer and FaceXHuBERT (Ours). Given an audio sequence from the test-set, \cref{fig:ff_ours_gt} is the ground-truth whereas \cref{fig:ff_ours_ff,fig:ff_ours_ours} are the corresponding frames from animations generated using FaceFormer and FaceXHuBERT respectively. \cref{fig:ff_ours_diff} shows the colorized per vertex distance computed between \cref{fig:ff_ours_ff,fig:ff_ours_ours}. Animation generated by our approach is more expressive because the upper face region is more responsive and expressive to the emotionally expressive audio sequences than the one of FaceFormer.}
  \label{fig:ff_ours}
  \vspace{-1em}
\end{figure}

The quality of animations generated by FaceXHuBERT has been studied carefully to understand the generalizability capabilities of the model in terms of various aspects such as- smoothness of the animation, coherence with respect to speech, different background noises in input audio, multiple speakers, different languages, different subjects, use of TTS (text2speech) instead of real audio and limited training data. The proposed approach is robust in terms of all the above-mentioned aspects. Additionally, by training the network without the binary emotion label (FaceXHuBERT- w/o emo) in the decoder, we lose the expressive style control capability during inference. In this case, facial expressiveness of the generated animations rely solely on the audio signal and are qualitatively slightly less expressive than the proposed model's predictions. Yet it still captures and distinguishes between neutral and emotional aspects in speech signals.  Furthermore, we qualitatively compared our approach to the FaceFormer (see \cref{fig:ff_ours}) and found that FaceXHuBERT generates more expressive animations that are closer to the ground-truth. Unlike our approach, FaceFormer is not robust to variety of noises overlapping with the audio signal and produces visual artefacts. We recommend watching the supplementary video for visual quality judgement.

\subsection{Perceptual Evaluation}
\label{sec:perceptual_evaluation}

\epigraph{``A work of art doesn’t exist outside the perception of the audience.''}{\textit{Abbas Kiarostami}}

In order to demonstrate the realism of facial animation produced by our proposed approach and to compare to ground-truth and current state-of-the-art, we conducted perceptual user studies. The user studies were hosted on Qualtrics \cite{qualtrics} and carried out using Prolific \cite{prolific}, ensuring that the participants get compensated appropriately. We conducted three separate user studies where the users selected their preferences based on the realism of the rendered 3D facial animation videos. In Experiment \RNum{1}, the users indicated their preference between Ground-truth data vs Ours. In Experiment \RNum{2}, the users were shown side-by-side comparisons of FaceFormer and FaceXHuBERT generated with the same audio input. Lastly, in Experiment \RNum{3} the users depicted their preference between neutrally and expressively generated animations both using our approach. \cref{fig:emoeffect} quantitatively and visually shows the difference between neutrally and expressively generated animations. The facial regions (i.e. upper face, eye region, cheeks) that are expected to be effected by the emotion condition deform differently than in the corresponding neutral animation. To solidify our argument, the third experiment was conducted to prove that the expressiveness in facial animation is actually perceived by users.

\begingroup
\setlength{\tabcolsep}{4pt} 
\renewcommand{\arraystretch}{0.6} 
\begin{table}[t]
  \centering
  \begin{tabular}{c c}
    \toprule
    User Preference & Realism\\
    \midrule
    \RNum{1}. Ours \textit{vs.} Ground-Truth &  $25.45\pm13.20$\\
    \RNum{2}. Ours \textit{vs.} FaceFormer &  $77.73\pm17.59$\\
    \RNum{3}. Ours-Emotional \textit{vs.} Ours-Neutral & $66.35\pm5.66$\\
    \bottomrule
  \end{tabular}
  \caption{\textbf{User study results:} The Realism score depicts the average percentage of participants that preferred the left item to the right item in the corresponding row in terms of expressiveness.}
  \label{tab:user_study}
  \vspace{-5mm}
\end{table}
\endgroup

For all three experiments, the participants where shown a series of A vs. B video pairs and were asked to choose one between the two animations that looks more realistic and more expressive. In order to ensure good quality data, we included random attention check questions. The attention check questions consist of a video pair in which one of the videos is totally out-of-sync with the accompanied audio. 

The results of the three experiments are reported in \cref{tab:user_study}. In Experiment \RNum{1}, on average, 25.45\% of the participants preferred animation generated by our model over the ground-truth animation. This is expected as the generated animations do not model some of the facial motion such as the eye blinks that are present in the ground-truth. Furthermore, the ground-truth has more variations across the whole face. In Experiment \RNum{2}, on an average, 77.73\% of the participants preferred animations generated by our model over the animations generated by FaceFormer. In Experiment \RNum{3}, on average, 66.35\% of the participants preferred animations generated by our model with expressive signal over the animations generated without the expressive signal.

\section{Ablation Study}
\paragraph{Ablation on FaceXHuBERT Encoder}
\label{sec:ablation_hubert}
We conducted extensive experiments to understand and optimize the effect of HuBERT in our encoder. This ablation study is devised by freezing the model's pretrained weights at various layers starting from no-freezing (i.e. all the layer parameters are trainable) to all weights frozen (i.e. all the layers are frozen to their pretrained weights). The results of the ablation study is reported in \cref{tab:Hubert_ablation}. In light of the encoder structure as described in \cref{sec:enc}, the network configurations for this ablation study are as follows- \textbf{(i)} no layers are frozen, \textbf{(ii)} CNN encoder is frozen, \textbf{(iii)} CNN encoder and feature projection layers are frozen. For models \textbf{(iv)} to \textbf{(viii)}, in addition to freezing the CNN encoder and feature projection layer, we incrementally freeze two transformer layers. For model \textbf{(ix)}, only the last transformer layer is kept trainable whereas for \textbf{(x)}, the entire pretrained model is frozen during training. Although all the above mentioned configurations yield qualitatively coherent animations, we found that \textbf{(iv)} produces the least mean face vertex error. Furthermore, as we move from model \textbf{(iv)} to \textbf{(x)}, the animations become more stiffer and less expressive. 
\vspace{-1.5em}
\paragraph{Ablation on FaceXHuBERT Decoder}
\label{sec:ablation_gru}
We experimented with different configurations of the GRU structure in terms of number of hidden layers and hidden unit size to optimize for the mean vertex error. \cref{tab:GRU_ablation} shows the results of the FaceXHuBERT decoder ablation study. The first column represents the number of hidden layers (e.g. 2L) and hidden size (e.g. 128) in the GRU structure. All the configurations mentioned in the table produce qualitatively coherent animations but training the proposed configuration results in producing the least mean face vertex error, ensuring that the predictions are more closer to the ground-truth, hence more realistic and expressive.

\section{Discussion and Limitations}
Using a self-supervised pretrained speech model such as HuBERT produces significant improvements for the 3D speech-driven facial animation task. It clearly shows the importance of the encoder model, while using a simple decoder component. It does not only have the ability to disambiguate speech uncorrelated factors for facial animation, but also addresses the scarcity of synchronized audio-visual datasets by incorporating pretrained speech representations based on a large speech model. We assume that it can be adopted to solve similar downstream tasks such as audio-driven gesture synthesis. Additionally, we showed that guiding the training with an emotion label captures the facial deformations uncorrelated with and correlated with emotion context. 

Due to the limitation of the BIWI dataset, we could only guide the learning in a binary manner (i.e. neutral and expressive). However, we assume that with a balanced dataset containing specific emotion categories in the data, we will be able to learn and generate audio-driven facial animations for respective emotion categories. Furthermore, our approach is limited to offline animation generation. In the future, we plan to extend our work to be real-time friendly. Our proposed approach works on 3D meshes and still needs to be mapped to a rigged character. Additionally, since the face scans of the dataset do not contain eyes and tongue, our method could not take into account animations of some face parts such as eye gaze and tongue. 
\begingroup
\setlength{\tabcolsep}{5pt} 
\renewcommand{\arraystretch}{0.6} 
\begin{table}[t]
  \centering
  \begin{tabular}{c c c c}
    \toprule
    Model & Mean Face & No. of  & Training\\
    variant & Vertex Error (mm) & parameters & Time (h)\\
    \midrule
    (i) & $5.17$ & $114166622$ & $\approx8.19$\\
    (ii) & $4.8$6 & $109966174$ & $\approx5.75$\\
    (iii) & $5.00$ & $109571166$ & $\approx6.38$\\
    \textbf{(iv)} & $4.80$ & $95395422$ & $\approx5.10$\\
    (v) & $4.84$ & $81219678$ & $\approx5.83$\\
    (vi) & $4.85$ & $67043934$ & $\approx5.50$\\
    (vii) & $4.89$ & $52868190$ & $\approx5.27$\\
    (viii) & $5.02$ & $38692446$ & $\approx5.13$\\
    (ix) & $4.95$ & $31604574$ & $\approx4.86$\\
    (x) & $5.24$ & $19795678$ & $\approx3.88$\\
    \bottomrule
  \end{tabular}
  \caption{FaceXHuBERT Encoder Ablation Study Results. Model \textbf{(iv)} depicts the proposed approach.}
  \label{tab:Hubert_ablation}
\end{table}
\endgroup

\begingroup
\setlength{\tabcolsep}{4pt} 
\renewcommand{\arraystretch}{0.6} 
\begin{table}[t]
  \centering
  \begin{tabular}{c c c c}
    \toprule
    Model & Mean Face  & No. of  & Training\\
    variant & Vertex Error (mm) & parameters & Time (h)\\
    \midrule
    \textbf{2L-256} & $\mathbf{4.80}$ & $95395422$ & $\approx5.10$\\
    1L-256 & $4.93$ & $95000670$ & $\approx5.55$\\
    2L-128 & $4.92$ & $85385566$ & $\approx5.33$\\
    2L-64 & $5.00$ & $80491230$ & $\approx5.27$\\
    2L-32 & $5.19$ & $78071710$ & $\approx4.30$\\
    \bottomrule
  \end{tabular}
  \caption{FaceXHuBERT Decoder Ablation Study Results. Model \textbf{2L-256} depicts the proposed approach.}
  \label{tab:GRU_ablation}
  \vspace{-1em}
\end{table}
\endgroup

\section{Conclusion}

In this paper, we presented FaceXHuBERT, an efficient text-less speech-driven expressive 3D facial animation method using self-supervised speech representation learning. At the core of our model is the pretrained HuBERT-based encoder combined with an efficient GRU-based decoder instead of a complex model based on transformers. Our method can produce accurate lip-sync and expressive facial animations for arbitrary audio input without the need of long training times and large dataset. Our method does not only produce accurate lip-sync but also captures personalized and subtle cues in speech (e.g. identity, emotion and hesitation). It is also very robust to background noise and can handle audio recorded in a variety of situations (e.g. multiple people speaking, background noise, laughter, lip-smacking). Our extensive objective and subjective analyses show that FaceXHuBERT outperforms the state-of-the-art. We hope that our approach will be a stepping stone towards text-less speech-driven expressive 3D facial animation.

\noindent\textbf{Ethical Consideration} Models trained on face scans can easily be used for generating synthetic content that can jeopardize humans and their privacy. We must act responsibly by considering the aspects pertaining to privacy and ethics.

\noindent\textbf{Acknowledgement} We would like to thank the authors of FaceFormer for making their code available and to ETH Zurich CVL for providing us access to the \textit{Biwi 3D Audiovisual Corpus}. We express our gratitude to our colleagues for providing different language audios and for the feedback.

{\small
\bibliographystyle{ieee_fullname}
\bibliography{egbib}
}

\clearpage
\appendix
\section{Supplementary Material}

\begin{figure*}[]
  \centering
  \includegraphics[width=0.9\linewidth]{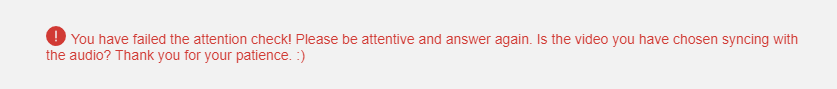}
  \caption{Attention check warning message.}
  \label{fig:userstudyscreen3}
\end{figure*}
\subsection{FaceXHuBERT Algorithm}
\cref{alg:training} depicts a high level pseudo code of the training procedure of FaceXHuBERT described in section \cref{sec:proposedapproach}. Given an audio waveform $A$, Subject Label, Emotion Label as inputs and corresponding 4D scan $Y$ as output, the proposed network learns the mapping between the inputs and the output during training epochs. The network optimizes on Huber Loss function and uses Adam optimizer to update the weights and biases during backpropagation. 

\begin{algorithm}

\caption{Network Training}\label{alg:training}
\begin{algorithmic}
\State Given input audio ($A$) and output 4D-Scan ($Y$):
\State $A = AudioWaveform$ \Comment{Raw Audio Waveform}
\State $Y = [T,V]$ \space -D Matrix \Comment{Ground-Truth 4D Data} 
\newline

\State $S \gets SubjectEmbedding(SubjectOneHot)$
\State $E \gets EmotionEmbedding(EmotionOneHot)$
\State $B \gets AudioEncoder(A)$
\State $X \gets InputRepresentationAdjustment(B)$
\State $H \gets GRU(X)$
\State $\Tilde{H} \gets H \odot S \odot E$
\State $\hat{Y} \gets FullyConnected(\Tilde{H})$
\State $\mathcal{L} \gets HuberLoss(Y, \hat{Y})$
\State $AdamOptimizer()$

\end{algorithmic}
\end{algorithm}

\subsection{Input Representation Adjustment}
This module adjusts the input representation and output representations and do not contain any trainable parameters. This is devised to ensure the one-to-one frame level relationship between input $X$ and output $Y$ such that $T_X = T_Y = T$. This function is generically devised in such a way that it can handle any input-output frequency pair given, $f_{o}\leq f_{i}$, where $f_{i}$ is the frequency of encoded discreet representation of the input audio and $f_{o}$ is the frequency of the face scan data. If $\frac{f_{i}}{f_{o}}=k \in Z^+$ where $Z^+$ denotes the set of positive integers, then the adjustment is a straightforward reshape function such that input dimension $(T_X,B) = (kT_Y,B)$ becomes a $(T_Y,kB)$ dimensional data. If $k \notin Z^+$, we resample the input representation using linear interpolation so that input dimension $(T_X,B)$ becomes $(\lceil k \rceil T_Y,B)$ before reshaping the embedding into $(T_Y,\lceil k \rceil B)$ dimensional data to ensure $T_X= T_Y = T$. Here $\lceil k \rceil$ denotes the ceiling of the decimal representation, in other words we take the next positive integer. In our implementation, we ensure that the value of $k = 2$ for it to be coherent with the implemented network architecture. In case of other datasets, where $k\neq2$, the model definition needs to be slightly modified in terms of input dimension of the GRU in the decoder.

\subsection{Data Pre-process}
\cref{alg:preprocess} represents the dataset pre-processing procedure. This step is a prerequisite to train the proposed model effectively. The vertex data in BIWI dataset is not normalized. Without scaling the data to a certain uniform range, the network does not train well. Although this is not the only way to normalize the data, we highly recommend normalizing the dataset so that all three coordinates (i.e. 3D coordinates- X, Y, Z) have the same range of values (e.g. [-0.5,0.5], [-1,1] or [0,1]) before starting training. Appropriate step-by-step guide together with code to prepare and pre-process the dataset are available in the project's GitHub repository provided with the supplementary material to facilitate reproducibility of our work.   

\begin{algorithm}
\caption{Data Pre-process}\label{alg:preprocess}
\begin{algorithmic}

\State $Neutral = [Array Of Subjects With Neutral Faces]$
\State $Templates = [EmptyArray]$

\While{There is $Subject$ to process}
\State $S \gets Neutral[Subject]$
\State $X \gets S[:, 0]$; $Y \gets S[:, 1]$; $Z \gets S[:, 2]$ 

\State $S[:, 0] \gets (X - mean(X)) \div (max(X) - min(X))$
\State $S[:, 1] \gets (Y - mean(Y)) \div (max(Y) - min(Y))$
\State $S[:, 2] \gets (Z - mean(Z)) \div (max(Z) - min(Z))$

\State $Templates.append(S)$

\While{There is $sequence$ to process}

\While{There is $frame$ to process}


\State $\Tilde{X} \gets \frac{(sequence[frame,:,0] - mean(X))}{(max(X) - min(X))}$
\newline
\State $\Tilde{Y} \gets \frac{(sequence[frame,:,1] - mean(Y))}{(max(Y) - min(Y))}$
\newline
\State $\Tilde{Z} \gets \frac{(sequence[frame,:,2] - mean(Z))}{(max(Z) - min(Z))}$

\State $sequence[frame,:,0] \gets \Tilde{X}$
\State $sequence[frame,:,1] \gets \Tilde{Y}$
\State $sequence[frame,:,2] \gets \Tilde{Z}$

\EndWhile

\State $SaveProcessedSequence(sequence)$

\EndWhile

\EndWhile

\end{algorithmic}
\end{algorithm}

\subsection{User studies}
\begin{figure}[h!]
  \centering
  \includegraphics[width=0.9\linewidth]{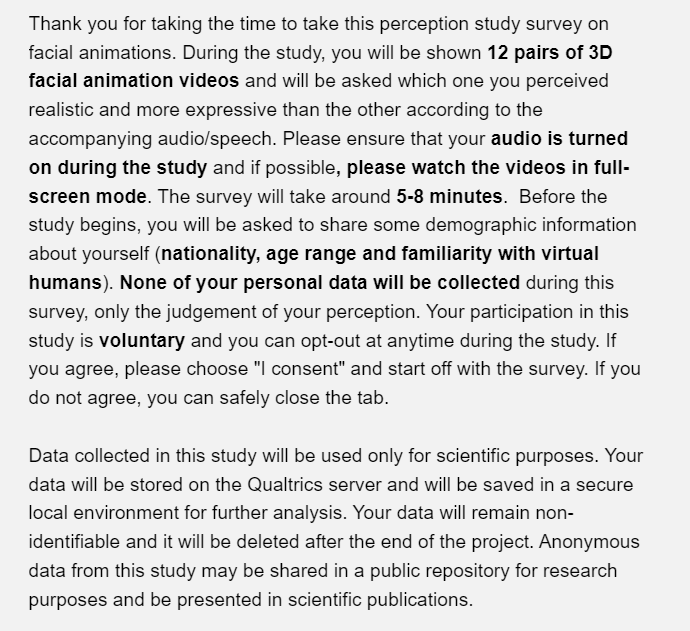}
  \caption{Introduction page of the user study surveys.}
  \label{fig:userstudyscreen1}
\end{figure}

\begin{figure}[h!]
  \centering
  \includegraphics[width=0.9\linewidth]{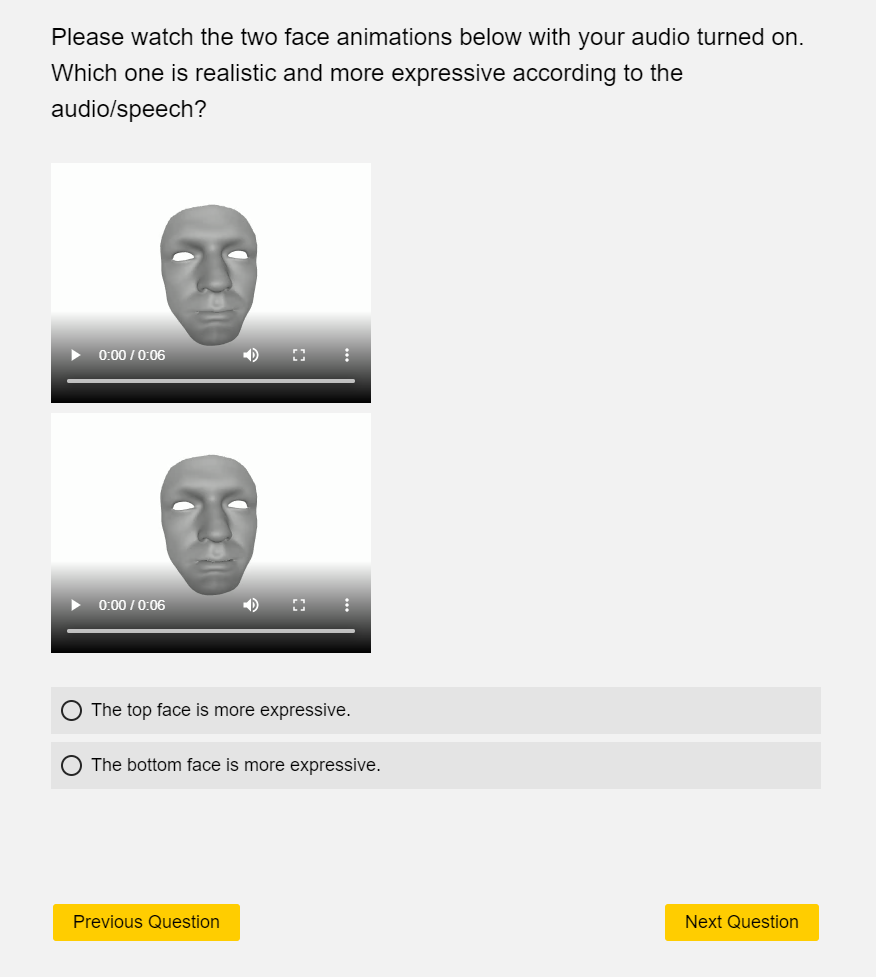}
  \caption{User interface of user study surveys.}
  \label{fig:userstudyscreen2}
\end{figure}

For perceptual evaluation described in section \cref{sec:perceptual_evaluation}, three similar user study experiments were conducted. In all three experiments, the participants were randomly shown 12 pairs (to ensure a good duration of the study) of facial animation videos and asked to choose the one that is realistic and more expressive than the other in accordance with the audio. \cref{fig:userstudyscreen1} is the introduction message that the participants were shown. \cref{fig:userstudyscreen2} shows the user interface of the survey related to the user study experiments. In total, we were able to recruit 147 participants from different demographic backgrounds for the experiments ensuring a good sample representation of the population. Among this 147 participants, 51 participated in Experiment \RNum{1}, 31 in Experiment \RNum{2} and 65 in Experiment \RNum{3}. 
\vspace{-1em}

Additionally, the participants were shown attention check question items, where one of the videos was facial animation that was generated by a model trained on MFCC (Mel-frequency cepstral coefficients) features instead of FaceXHuBERT Encoder and the other is either ground-truth or generated by our approach. The animations generated by MFCC based model do not produce coherent animations where the lip-sync is incongruous to the accompanied audio. If the participants had chosen the animation video generated by MFCC based model, there were shown the warning message depicted in \cref{fig:userstudyscreen3}. Those participants' responses are then manually reviewed for inconsistency in the user study data.

\end{document}